\documentclass[journal]{IEEEtran}
\ifCLASSINFOpdf
\else
\fi
\usepackage{graphicx}
\usepackage{amssymb}
\usepackage{subfigure}
\usepackage{cite}
\usepackage{color}
\usepackage{amsmath,amsfonts,algorithmic}

\begin{document}
%
% paper title
% Titles are generally capitalized except for words such as a, an, and, as,
% at, but, by, for, in, nor, of, on, or, the, to and up, which are usually
% not capitalized unless they are the first or last word of the title.
% Linebreaks \\ can be used within to get better formatting as desired.
% Do not put math or special symbols in the title.
\title{Sensor Drift Compensation in Electronic-Nose-Based Gas Recognition Using Knowledge Distillation}
%
%
% author names and IEEE memberships
% note positions of commas and nonbreaking spaces ( ~ ) LaTeX will not break
% a structure at a ~ so this keeps an author's name from being broken across
% two lines.
% use \thanks{} to gain access to the first footnote area
% a separate \thanks must be used for each paragraph as LaTeX2e's \thanks
% was not built to handle multiple paragraphs
%

\author{Juntao Lin, Xianghao Zhan, \IEEEmembership{Member, IEEE}
\thanks{Submitted date: . This work was not founded by any specific research grant.}
\thanks{Juntao Lin is with School of Zhejiang University, College of Biosystems Engineering and Food Science, Hangzhou, Zhejiang, 310000, China (e-mail: 3200101629@zju.edu.cn).}
\thanks{Xianghao Zhan, Ph.D., is with Department of Bioengineering, Stanford University, Stanford, CA, 94305, USA (e-mail: xzhan96@stanford.edu).}
\thanks{Corresponding author: Xianghao Zhan}}

\maketitle

% As a general rule, do not put math, special symbols or citations
% in the abstract or keywords.
\begin{abstract}
%% Text of abstract
Objectives: Due to environmental changes and sensor aging, sensor drift challenges the performance of electronic nose systems in gas classification in real-world deployment. Previous studies using the UCI Gas Sensor Array Drift Dataset reported promising drift compensation results but lacked robust statistical experimental validation and may overcompensate sensor drift and lose class-related variance. Methods: to address these limitations and explore improvement in sensor drift compensation with statistical rigor, we firstly designed two domain adaptation tasks based on the same electronic nose dataset: 1) using the first batch to predict remaining batches, simulating a controlled laboratory setting, and 2) predicting the next batch using all prior batches, simulating continuous training data updates for online training. Then, we systematically tested three methods: our proposed novel Knowledge Distillation method (KD), the benchmark method Domain Regularized Component Analysis (DRCA), and a hybrid method KD-DRCA, across 30 random test set partitions on the UCI dataset. Results: we showed that KD consistently outperformed both DRCA and KD-DRCA, achieving up to 18\% improvement in accuracy and 15\% in F1-score, proving KD’s superior effectiveness in drift compensation. Conclusion: This is the first application of KD for electronic nose drift mitigation, which significantly outperformed the previous state-of-the-art drift compensation DRCA method, enhancing the reliability of sensor drift compensation in real-world environments.
\end{abstract}

% Note that keywords are not normally used for peerreview papers.

% For peer review papers, you can put extra information on the cover
% page as needed:
% \ifCLASSOPTIONpeerreview
% \begin{center} \bfseries EDICS Category: 3-BBND \end{center}
% \fi
%
% For peerreview papers, this IEEEtran command inserts a page break and
% creates the second title. It will be ignored for other modes.
\IEEEpeerreviewmaketitle

\section{Introduction}
\label{sec:introduction}
%用电子鼻分类gas的主要意义
\IEEEPARstart{N}{on}-intrusive gas recognition is significant in research and industrial applications \cite{bhattacharyya2008electronic, sarkodie2018review}. Electronic nose, an artificial olfactory system that mimics the mammalian olfactory system, offers a rapid, cost-effective, and non-intrusive method for analyzing different types of volatile organic compounds and gases \cite{wang2019discrimination,zhan2018discrimination}. Its application in gas classification is simple and cost-effective \cite{zhan2020electronic, borowik2021application}. The easy operation and low cost enhance its versatility and application fields \cite{flammini2004low}. Accurate identification and classification of gases with an electronic nose have numerous applications across fields\cite{pace2016developments}: in environment monitoring, electronic noses have been used to assess air quality \cite{de2009co}; in food quality assessment, electronic noses can detect the gas composition of food as a way to determine the quality of food \cite{musatov2010assessment}. For example, Wojnowski's team used electronic noses to successfully classify poultry and canola oil samples and detect olive oil adulteration with an overall accuracy of 82\% \cite{wojnowski2017portable}. In addition, electronic noses are used to detect the quality of beverages and medicinal herbs such as wine, coffee, ginseng and Chinese medicine \cite{loutfi2015electronic,rodriguez2009electronic, miao2016comparison,zhan2019feature}. Furthermore, there are applications to aid in disease diagnosis by analyzing volatile organic compounds in air \cite{schmekel2014analysis,zhan2020electronic}. On the pattern recognition software and algorithm side, in the past two decades, with the development of machine learning, integrating pattern recognition algorithms, such as support vector machines, neural networks, and conformal prediction, has enabled electronic noses to perform accurate classification and regression analysis of various analytes \cite{rodriguez2014calibration,jha2014neural,zhan2018online}. For example, the electronic nose developed by Brudzewski et al. has been successfully used for the recognition of coffee with machine learning \cite{brudzewski2012recognition}. 

%2. 为什么存在域迁移的问题
Although the electronic nose is a powerful tool for gas classification, it faces a critical challenge of sensor drift in deployment. Sensor drift can result from factors such as aging of sensor materials, poisoning, and accumulation of contaminants. Furthermore, environmental changes, including variations in humidity and temperature, can also contribute to sensor drift \cite{hossein2010compensation}. These changes lead to variations in sensor response characteristics, which affects the robustness and accuracy of electronic nose-based pattern recognition models over time or across environments \cite{wang2019discrimination,zhan2018online}. Furthermore, sensor drift shows non-linear dynamic behavior in multi-dimensional sensor arrays, making it a challenging issue to address experimentally and computationally\cite{zhang2013chaotic}.

% 解决域迁移问题的现有方法
Sensor drift can be generally categorized as a domain adaptation problem in machine learning as sensor drift, which resembles the distribution drifts from source domain to target domain. Although transfer learning and model fine-tuning can be used to adapt the model to new data distributions \cite{zhan2024brain}, it requires supervised fine-tuning and labels on the new domains, which are hard to get in the electronic nose applications. This calls for unsupervised methods for drift compensation. The existing methods to address sensor drift can be categorized from a data perspective or from a model perspective. On the one hand, from the data perspective, domain adaptation methods are employed to align the data distributions between the target and source domains. As an example, Domain Regularized Component Analysis (DRCA) seeks to mitigate the differences between the target and source domains by finding a domain-invariant feature subspace. By mitigating these differences, DRCA can improve the generalizability of the model in the target domain. Previous studies have demonstrated that DRCA can effectively compensate for sensor drift in electronic nose, inertial measurement units (IMUs) and surface electromyography (sEMG) applications, yielding promising results in the target domain \cite{zhang2017anti,wang2022unsupervised,zhan2024adaptive}. Another data-perspective domain adaptation method, cycle-generative adversarial network (Cycle-GAN), transforms target domain data into source domain data, allowing the source domain models to be used for target domain data classification\cite{zhan2024adaptive}. However, the complex Cycle-GAN generally suffers from overfitting and a previous study directly comparing Cycle-GAN with DRCA showed its inferior performance \cite{zhan2024adaptive}.

On the other hand, from the model perspective, domain adaptation methods adapt the source-domain model to the target domain data distribution. As an example, Knowledge Distillation (KD) is a recently developed semi-supervised domain adaptation approach that improves model generalizability by transferring knowledge from a complex model (teacher model) to a simpler model (student model), preventing overreliance on the source domain and enabling better performance in the target domain \cite{orbes2019knowledge}. Additionally, other semi-supervised learning approaches, such as reliability-based unlabeled data augmentation and training data curation using conformal prediction, involve attaching target domain data with pseudo-labels and expanding the training set based on the reliability of these labels to enhance model adaptability\cite{liu2021boost,liu2022cpsc,zhan2025reliability}. So far, KD has not been used in electronic nose-based applications yet, while the effectiveness of the semi-supervised reliability-based unlabeled data augmentation has been systematically and statistically validated in electronic nose, yet only in artificially simulated sensor drift \cite{liu2021boost,liu2022cpsc}.

Although these methods can improve generalization to some extent for specific tasks, each method has its limitations. For example, DRCA and Cycle-GAN can easily overfit samples from a particular domain, overcompensate domain drift and lose class-related variance, and therefore perform sub-optimally on classifying unseen target domain data, while KD and reliability-based unlabeled data augmentation do not directly address domain differences like DRCA and Cycle-GAN, which could lead to underfitting of the models for target domain usage \cite{liu2021boost,liu2022cpsc}. To be specific, the KD and reliability-based unlabeled data augmentation approaches may carry too much modeling inertia from the supervised portion from the source domain and therefore insufficiently address the domain drift \cite{liu2021boost,liu2022cpsc}.

To investigate the sensor drift for electronic-nose-based gas recognition, the UCI Gas Sensor Array Drift Dataset provides an ideal testbed with 10 batches of electronic-nose gas datasets \cite{misc_gas_sensor_array_drift_dataset_224}. Previously, there have been studies trying to develop and evaluate new drift compensation methods for electronic-nose-based gas recognition \cite{vergara2012chemical,chang2023study,dennler2022drift,zhang2017anti}. Particularly, DRCA has been set as a benchmark method for e-nose drift compensation on the dataset\cite{zhang2017anti}. However, the effectiveness of these domain adaptation approaches lack a statistically sound validation to address the sensor drift problem for electronic noses \cite{liu2021boost,zhang2017anti}. For example, the DRCA benchmark on the electronic-nose-based gas identification tasks has not been rigorously tested for statistical significance under stochasticity and randomized training disturbances. The previous studies to compensate electronic nose sensor drift only reported one-time test accuracy, and other important classification metrics like precision, recall, F1-score remains to be tested \cite{zhang2017anti,vergara2012chemical,chang2023study,rehman2020multi,dennler2022drift}. Additionally, the results still needs to be systematically validated under diverse tasks simulating the real-world application scenarios with parallel experiments and statistical significance tests\cite{liu2021boost}. Despite the increasing accuracy of gas classification systems, prior studies often report results based on a single test without statistical validation, leading to potentially over-optimistic performance that may not generalize across real-world applications. Additionally, there is a lack of methods that effectively combine both data and model perspectives to leverage their complementary strengths.

%本文的研究内容
To address the limitations of existing methods, to fill in the missing gaps in the addressing the sensor drift in electronic-nose-based pattern recognition, and to potentially improve the effectiveness of domain adaptation, in this study, we made our unique contributions: 

\begin{itemize}
    \item We designed a new systematic and statistical experiment protocol for the development and evaluation of drift compensation methods based on the public UCI dataset. We designed two realistic domain adaptation tasks: 1) predicting the remaining batches using the first batch, and 2) predicting the next batch using all previous batches. Task 1 simulates a well-controlled laboratory environment for model development, while Task 2 mimics a continuously updated training dataset for better online model training. Most importantly, unlike the previous studies reporting accuracy in a single test, statistical significance was tested with 30 random test set partitions for accuracy, precision, recall, F1-score, to systematically and statistically validate a method's robust performance under various sensor drift conditions. 
    \item Based on the refined experiment protocol, we tested the KD method (a new method not yet applied in sensor drift compensation in previous studies as far as we know) and the benchmark DRCA method (applied in sensor drift compensation in electronic nose \cite{zhang2017anti} but the validity was not systematically tested with rigorous statistical tests in experiments that better simulate real-world application scenarios) with various cross-domain prediction tasks and rigorous statistical tests. 
    \item We explored a novel hybrid approach that combines two domain adaptation strategies, using the electronic-nose-based gas classification task as a test case.
\end{itemize}

As a summary, based on the UCI Gas Sensor Array Drift Dataset, we conducted a series of cross-domain classification experiments under sensor drift conditions to simulate the real-world scenarios of using electronic nose for gas classification, and to systematically and statistically validate the effectiveness of the proposed KD method. It should be noted that, DRCA, which has been previously validated for its effectiveness in drift compensation \cite{zhang2017anti}, was used as both the baseline and benchmark for evaluating the KD method. Therefore, we regard DRCA as a reliable baseline approach in this study to be compared against.

% The configuration of domain adaptation enables the application of MLHMs to a new type of head impact without sacrificing accuracy nor developing a new MLHM with supervised learning, i.e., model finetuning, by running a large amount of FE simulations. This work will accelerate the application of MLHMs in the various head impact scenarios and ultimately contribute to the improvement of TBI diagnosis and protection.

% This paragraph is not used.

\section{Methods}

\subsection{Dataset Description}
This study utilizes the well-established Gas Sensor Array Drift Dataset from the University of California, Irvine (UCI) Machine Learning Repository \cite{misc_gas_sensor_array_drift_dataset_224}. This comprehensive dataset is a well-established benchmark designed to investigate the effects of sensor drift in gas classification. It includes measurements from 16 chemical sensors exposed to six different gases over 36 months. The classes of gases in the dataset are ammonia, acetaldehyde, acetone, ethylene, ethanol, and toluene. The primary objective of the dataset is to provide a robust foundation for developing and evaluating algorithms that mitigate the effects of sensor drift, a critical issue in the long-term deployment of electronic nose.

To investigate the sensor drift, the dataset is divided into ten batches, corresponding to different times of data collection. This structure enables a detailed analysis of the temporal variations in sensor responses for a better understanding of sensor drift and the development of effective drift compensation algorithms \cite{misc_gas_sensor_array_drift_dataset_224}.

%Although on this public dataset, previous methods have shown effectiveness in drift compensation \cite{vergara2012chemical,chang2023study,rehman2020multi,dennler2022drift}, they were all based on single test set partitioning and single-test protocol, which limits the ability to demonstrate the robustness and reproducibility of their results and avoid serendipity.

In this study, we use the term "source domain" to denote the data used for model development, and "target domain" to refer to the validation/test data under sensor drift conditions. Given the sequential collection of the ten batches (i.e., batch 10 was collected after batch 9), we developed two types of cross-domain prediction tasks:
\begin{itemize}
    \item \textbf{Task 1}: use the first batch as the source domain to predict the remaining batches (target domain)
    \item \textbf{Task 2}: use the first $n - 1, n = 2, ..., 10$  batches as the source domain to predict the $n$-th batch (target domain).
\end{itemize}

Task 1 allows the source domain to simulate a dataset collected in a specific well-controlled laboratory environment at a particular times for gas classification model development. Task 2 simulates the continuously updated training dataset for better model development. Meanwhile, the target domain simulates a real-world user environment where the developed electronic-nose-based gas recognition models are applied later in its product lifespan, accommodating the effects of sensor drift.

\subsection{Knowledge Distillation}

Knowledge Distillation (KD), originally aiming to reduce neural network complexity by transforming high-parameter models into simpler ones \cite{hinton2015distilling}, is novelly adapted for a domain-adaptation purpose in this study. The process involves training a student network to emulate a teacher network using a loss function, the distillation loss. In the context of domain adaptation for sensor drift compensation, we employ a teacher-student learning strategy: the teacher model is trained on source domain data in a supervised manner. It then generates soft labels for a combination of source and target data (i.e., the predicted probability distributions), which are subsequently used to train the student model. This approach leverages the unlabeled data from target domain. Distillation loss provides a soft representation of one-hot encoded labels, facilitating the mitigation of overfitting to the source domain data, smoother optimization and the learning of label similarities \cite{orbes2019knowledge}. 

\subsubsection{Training the Teacher Model with Source Domain Data}
Consider a set of \( N(S) \) annotated samples from a source domain \( \mathbf{X}_S = \{(x_i, y_i), i = 1, \ldots, N(S)\} \), where \( x_i \in \mathbb{R}^d \) represents a \( d \)-dimensional feature, and \( y_i \in \{0, 1\}^K \) is its corresponding label in one-hot coding. Assuming there is a set \( \mathcal{F}_S \) that holds functions \( f : \mathbb{R}^d \rightarrow \mathbb{R}^K \), we firstly aim to learn a feature representation \( f_S \) (teacher model) via the minimization of a loss function, \( l \), according to Eq. (\ref{eq1}).

\begin{equation}
\label{eq1}
\underset{f \in \mathcal{F}_S}{\mathrm{arg \, min}} \frac{1}{N(S)} \sum_{x_i \in \mathbf{X}_S} l(y_i, \sigma(f_S(x_i)))
\end{equation}

\begin{equation}
\label{eq2}
[\sigma(z)]_k = \frac{e^{[z]_k}}{\sum_{l=1}^K e^{[z]_l}}
\end{equation}

Similar to standard supervised learning, the teacher network is optimized using the cross-entropy loss function.

\subsubsection{Training the Student Model for Target Domain}
Even if \( f_S \) is suitable to classify the measurements from the source domain \( \mathbf{X}_S \), it may not be suitable for data coming from a different testing distribution \( \mathbf{X}_T \). We aim to find another function \( f_T \in \mathcal{F}_T \), which is suitable to classify data from \( \mathbf{X}_T \). Assuming we have access to a limited set of unlabeled samples in the target domain \( \mathbf{X}_T = \{x_i, i = 1, \ldots, N(T)\} \), we can create a set
\begin{equation}
\begin{aligned}
X_U = \{(x_i, y_i) \mid x_i \in \mathbf{X}_S, y_i = f_S(x_i), 1 \leq i \leq N(S)\} \cup \\
\{(x_i, y_i) \mid x_i \in \mathbf{X}_T, y_i = f_S(x_i), 1 \leq i \leq N(T)\}
\end{aligned}
\end{equation}
that is used to optimize a student using the distillation loss. Through soft labels of this union $X_U$, the student is expected to learn a better mapping to the labels than the teacher network by mitigating the overfitting to the source-domain data. When training the student network, we consider probability distributions over the labels as targets. This representation reflects the uncertainty of the prediction by the teacher network. The function \( f_T \) is found by (approximately) solving,
\begin{equation}
\label{eq3}
\underset{f \in \mathcal{F}_T}{\mathrm{arg \, min}} \frac{1}{N(S) + N(T)} \sum_{x_i \in X_U} l(\sigma(T^{-1} f_S(x_i)), \sigma(f_T(x_i)))
\end{equation}
Here, \( T > 1 \) is the temperature parameter which controls the smoothness of the class probability prediction given by \( f_S \).

\begin{figure}[!ht]
	\centering
	\includegraphics[width=0.95\linewidth]{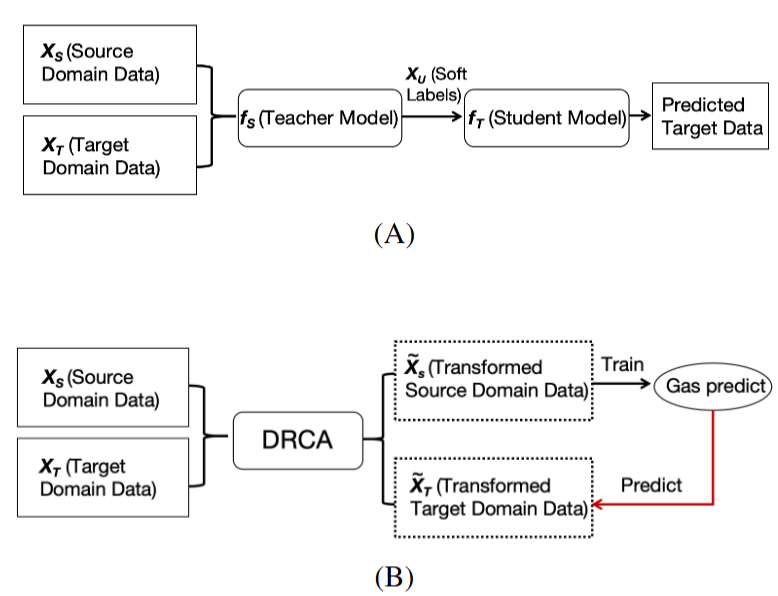}
	\caption{\textbf{Pipeline of KD and DRCA for drift compensation.} (A) The process involves training a teacher model on source domain to generate soft labels, which trains the student model. (B) DRCA transforms the source and target domain data into a new feature subspace. The transformed source data is used to train the model to predict the transformed target domain data.}
	\label{fig:processs}
\end{figure}

\subsection{Domain Regularized Component Analysis}
Domain Regularized Component Analysis (DRCA) \cite{zhang2017anti,wang2022unsupervised}, a representative of a domain adaptation approach from the data perspective, transforms features into a subspace through a linear projection that minimizes scatter between source and target domains while preserving within-domain scatter. Specifically, DRCA \cite{zhang2017anti} projects features to a hyper-plane that minimizes scatter between the source domain and target domain post-projection. Let the data in the source domain be denoted as \( x_i^S \in \mathbb{R}^{D} \), where \( i=1,2,...,N(S) \), and \( N(S) \) represents the number of samples in the source domain. Similarly, samples in the target domain are denoted as \( x_i^T \in \mathbb{R}^{D} \), where \( i=1,2,...,N(T) \), and \( D \) denotes the dimension of the original feature space\cite{zhan2024adaptive}.

Summary statistics of data distribution in the original feature space include:

a) Mean of source domain measurements: 
\[
\mu^S = \frac{ \sum_{i=1}^{N(S)}x_i^S}{N(S)} \in \mathbb{R}^{D}
\]
mean of target domain measurements:
\[
\mu^T = \frac{\sum_{i=1}^{N(T)}x_i^T}{N(T)} \in \mathbb{R}^{D}
\]
and overall mean:
\[
\mu = \frac{N(S) \times \mu^S + N(T) \times \mu^T}{N(S)+N(T)}
\]

b) Within-domain scatter for source and target domains:
\[
S_w^S = \sum_{i=1}^{N(S)}(x_i^S-\mu^S)(x_i^S-\mu^S)' \in \mathbb{R}^{D \times D}
\]
\[
S_w^T = \sum_{i=1}^{N(T)}(x_i^T-\mu^T)(x_i^T-\mu^T)' \in \mathbb{R}^{D \times D}.
\]

c) Between-domain scatter:
\[
S_b = (\mu^S - \mu^T)(\mu^S - \mu^T)' \in \mathbb{R}^{D \times D}.
\]

The optimization objective of DRCA is to find a projection matrix \( P \in \mathbb{R}^{D \times d}~(d<D) \) that minimizes between-domain scatter while maintaining within-domain scatter on the projected hyper-plane. Here, \( d \) is the dimension of the new feature space after projection. For a measurement \( x_i \) projected to \( \tilde{x}_i \): \( \tilde{x}_i = P'x_i \in \mathbb{R}^{d} \), summary statistics on the projected space include:
1. Within-domain scatter for projected source and target domain measurements: \( \tilde{S}_w^S = P'S_w^S P \in \mathbb{R}^{d \times d} \) and \( \tilde{S}_w^T = P'S_w^T P \in \mathbb{R}^{d \times d} \).
2. Between-domain scatter on the projection hyper-plane: \( \tilde{S}_b = P'S_b P \in \mathbb{R}^{d \times d} \).

The problem is formulated to maximize:
\begin{equation}
    \mathrm{max}_P \frac{\mathrm{tr}(P' S_w^S P + \alpha P' S_w^T P)}{\mathrm{tr}(P' S_b P)},
\end{equation}
where \( \alpha \) is a hyperparameter. By introducing a Lagrangian multiplier, the Lagrangian is expressed as:
\begin{equation}
    L(P, \theta) = \mathrm{tr}(P^T S_w^S P + \alpha P^T S_w^T P) - \theta (\mathrm{tr}(P^T S_b P) - \lambda),
\end{equation}
where \( \theta \) is the Lagrange multiplier and \( \lambda \) is a constant. Taking the derivative with respect to \( P \) and setting it to zero transforms the problem to an eigenvalue decomposition: 
\begin{equation}
S_b^{-1}(S_w^S + \alpha S_w^T) P = \theta P.
\end{equation}
 Sorting eigenvectors by eigenvalues, the top \( d \) eigenvectors form the projection matrix. This projects data from \( N(S) \times D \) and \( N(T) \times D \) to \( N(S) \times d \) and \( N(T) \times d \).

\subsection{Hybrid Method: KD-DRCA}
To address the limitations of existing methods and improve the effectiveness of domain adaptation, we explored to integrate DRCA and KD, the domain adaptation approaches from two different perspective. First, we use DRCA to project the data onto a feature subspace to mitigate the differences between the target and source domains. Subsequently, we used the KD to train a teacher model and student model on the projected data. Built upon the mitigated domain difference, KD further transfers knowledge from a complex model (teacher model) to a simpler model (student model), enabling the student model to generalize better in the target domain\cite{orbes2019knowledge}. We validate the effectiveness of the proposed method through a series of experiments.

\subsection{Model Development and Evaluation}
A Fully Connected Neural Network (FCNN) with four layers was used for the prediction tasks. The hidden units are: [100, 50, 20], with ReLU activation function, and a final output layer using the Softmax activation function for 6 classes. This structure was optimized as we firstly tested the within-batch classification with 70\%/15\%/15\% training, validation and test set and therefore, we kept this optimized model structure for cross-domain classification. Then, for the two cross-domain classification tasks, we randomly partitioned the target batch into 50\% validation and 50\% test data, with 30 random partitions to evaluate the model's robustness under randomness.

The following hyperparameters were tuned: the temperature parameter $T$ for KD (range: [0.3, 1, 2, 3, 5, 25, 50, 100, 200]), the subspace dimensionality $d$ (range: [50, 100]), and $\alpha$ (range: [0.001, 0.01, 0.1, 1, 10, 100, 1000]). A grid search was conducted to optimize accuracy on the validation set. 

On the validation and test sets, the performance of different methods across all tasks is evaluated using four metrics: accuracy, F1 score, recall, and precision, to derive comprehensive conclusions. The within-batch classification performance metrics were also recorded as the ideal performance without any domain drift.

% \begin{figure*}[!ht]
%     \centering
%     \includegraphics[width=0.99\linewidth]{bibtex/fig/Fig.1.pdf}
%     \caption{Distribution of MPS and MPSR of HM, CF1, MMA, CF2 and Boxing datasets. The distribution of the peak angular velocity magnitude (peak Ang. Vel. Mag., A), the peak angular acceleration magnitude (peak Ang. Acc. Mag. B), the 95th percentile MPS (C) and the distribution of the 95th percentile MPSR (D). The principal component analysis (PCA) visualization (E) and t-distributed stochastic neighbor embedding (t-SNE) visualization of all types of head impacts (F). Note: 95th percentile is frequently used in TBI biomechanics research to avoid numerical instability in the maximum values.}
%     \label{fig:1}
% \end{figure*}

\subsection{Statistical Test}
To test whether there are statistical significant differences between methods, paired t-tests were conducted on the results of 30 parallel experiments. 

%%%%%%%%%%%%%%%%%%%%%%%%%%
\section{Results}
\subsection{Data Visualization Under Sensor Drift}
\begin{figure*}[!ht]
    \centering
    \includegraphics[width=0.7\linewidth]{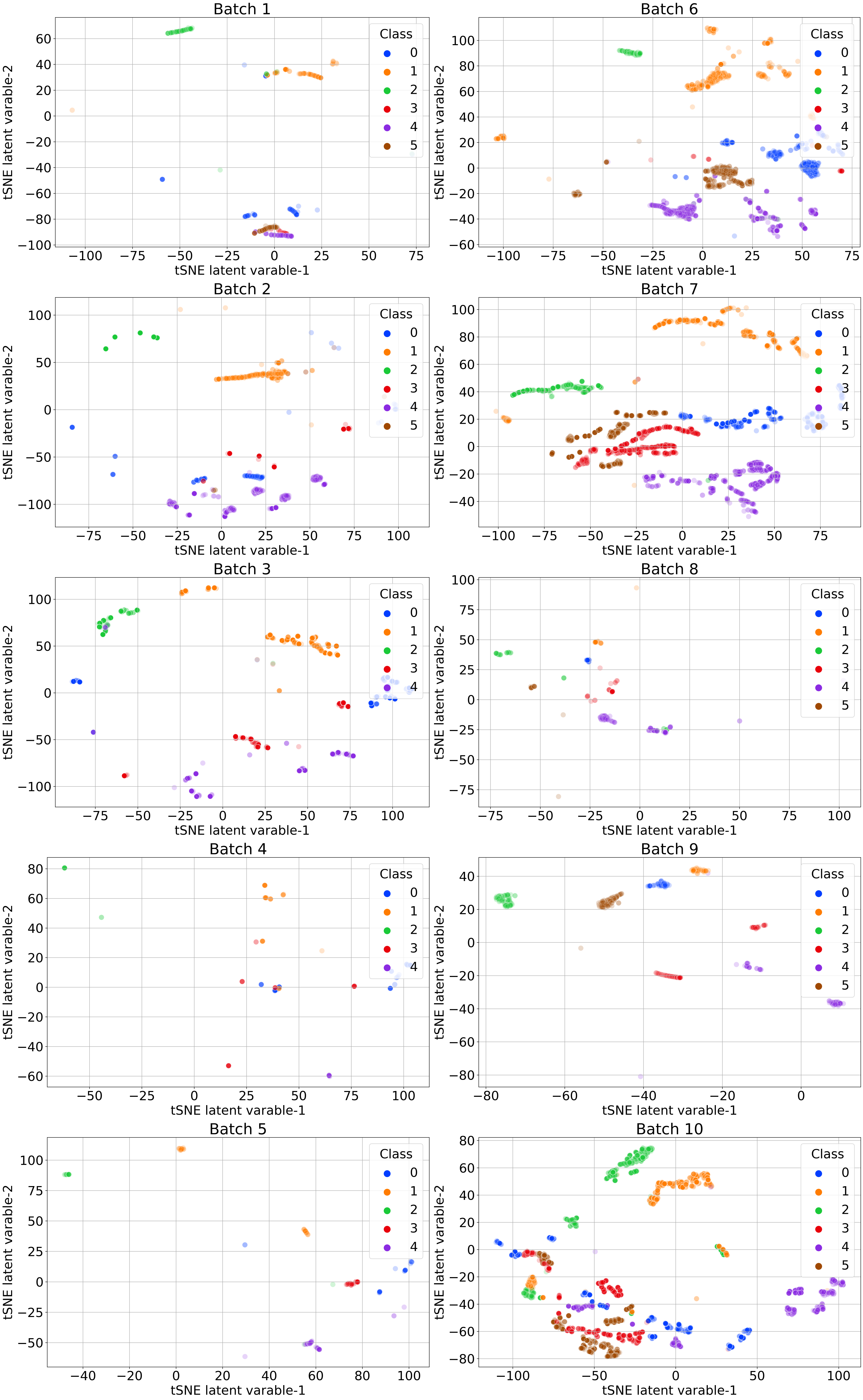}
    \caption{\textbf{The sensor drift visualization in the feature space of the ten batches of data with tSNE.} Classes: 0-Ethanol, 1-Ethylene, 2-Ammonia, 3-Acetaldehyde, 4-Acetone, 5-Toluene}
    \label{fig:tSNE}
\end{figure*}

We firstly visualized the feature-space data distribution of the ten batches in Fig. \ref{fig:tSNE}. The results show that clusters of different classes drift in the feature space across batches, which shows the effect of sensor drift over data collection times.

\subsection{Performance of Drift Compensation Methods}
\begin{figure*}[!ht]
    \centering
    \includegraphics[width=0.85\linewidth]{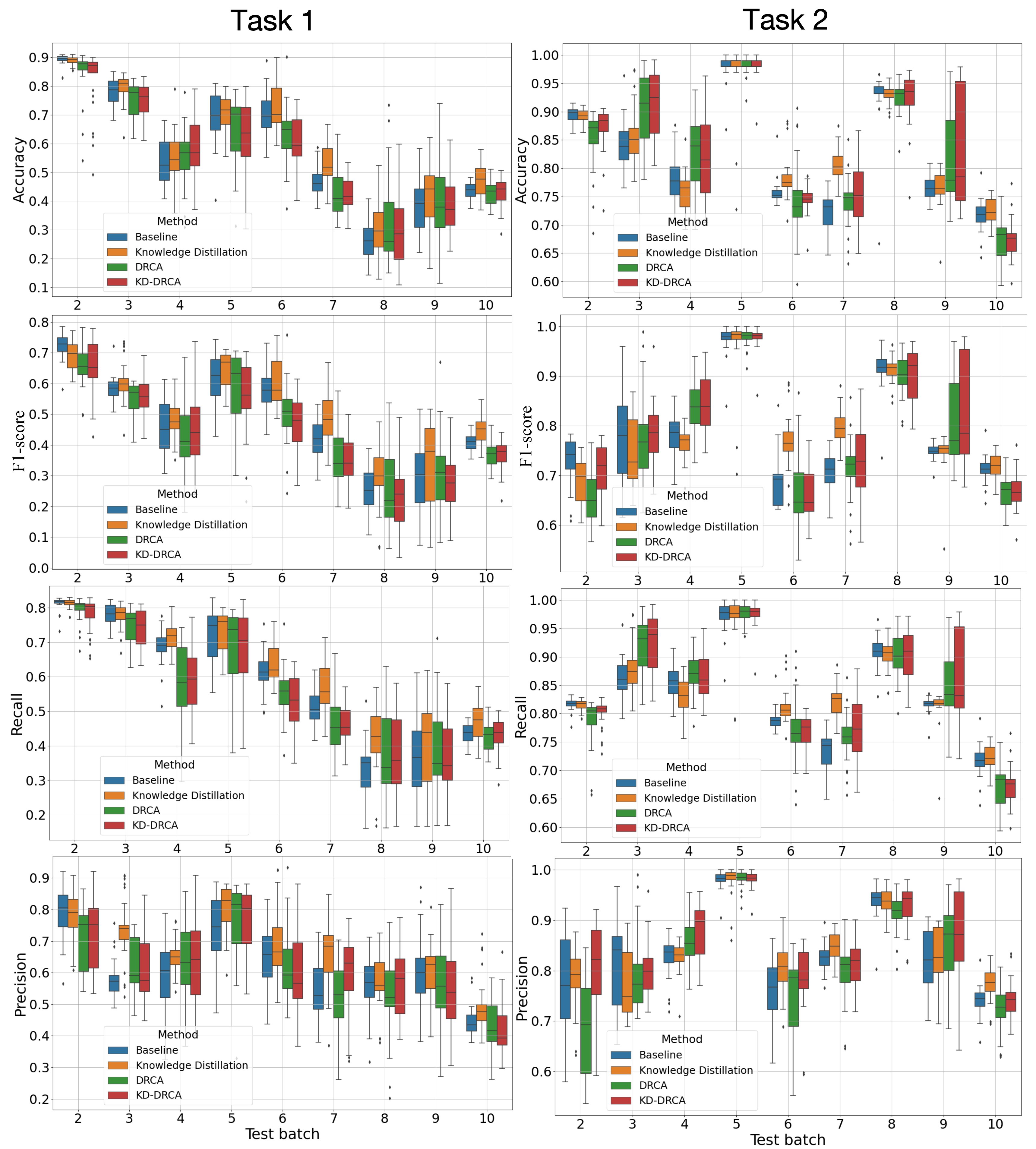}
    \caption{\textbf{Classification performance with the baseline, KD, DRCA, KD-DRCA methods on the test sets of different batches of data.} Task 1: Train on the first batch and test on the other batches; Task 2: Train on all previous batches and test on the next batch.}
    \label{fig:Boxplot}
\end{figure*}

With hyperparameter tuned, on the test sets, the classification performance are presented in Fig. \ref{fig:Boxplot}. It can be observed that on the test sets across batches, the KD method generally outperforms the Baseline, DRCA, and KD-DRCA methods in most cases. In task 1, over time (from batch 2 to batch 10), the overall classification performances with batch 1 as the source domain and any other batch as the target domain are decreasing, which clearly shows the gradually worsening sensor drift. Results on the validation sets are shown in Fig. S1. For most experiments, there is at least one drift compensation methods that outperforms the baseline. As a reference for the ideal classification performance without any sensor drift, the mean and standard deviation of the classification accuracy, F1-score, recall, precision across 10 within-batch classification tasks are: 0.979(0.035), 0.978(0.039), 0.979(0.035), 0.981(0.037). It can be seen that under sensor drift, the performances are generally worse than those on the within-batch classification tasks.

As a summary of cross-domain classification, Table 1 presents the statistical test results comparing all test experiments and metrics to the baseline. Table 2 presents the results on the test sets for task 1 and task 2 separately.

\begin{table}[h]
\centering
\caption{\textsc{The counts of statistical significant difference of the three methods comparing against the baseline. Each method has 72 samples: two cross-domain classification tasks, nine test batches for each task, four metrics. Inside the parentheses are the results on validation sets.}}
\begin{tabular}{lccccl}
\hline
\textbf{Method} & \textbf{+($p<0.05)$} & \textbf{=($p>0.05$)} & \textbf{-($p<0.05$)} & \textbf{Total} \\ 
\hline
\textbf{KD} & 24 (22) & 44 (48) & 4 (2) & 72 (72) \\
\textbf{DRCA} & 11 (12) & 37 (36) & 24 (24) & 72 (72) \\
\textbf{KD-DRCA} & 12 (12) & 38 (36) & 22 (24) & 72 (72) \\
\hline
\end{tabular}
\end{table}
\begin{table}[h]
\centering
\caption{\textsc{The counts of statistical significant difference on the three methods comparing against the baseline on the test sets in task 1 (left of "/") and task 2 (right of "/").}}
\begin{tabular}{lccccl}
\hline
\textbf{Method} & \textbf{+($p<0.05)$} & \textbf{=($p>0.05$)} & \textbf{-($p<0.05$)} & \textbf{Total} \\ 
\hline
\textbf{KD} & 15/9 & 20/24 & 1/3 & 36/36 \\
\textbf{DRCA} & 1/10 & 20/17 & 15/9 & 36/36 \\
\textbf{KD-DRCA} & 1/11 & 20/18 & 15/7 & 36/36 \\
\hline
\end{tabular}
\end{table}

Table 1 shows that both KD and KD-DRCA lead to more significantly improvement results ($p<0.05$) than DRCA, while KD yields the fewest significantly decrease results ($p<0.05$). Table 2 shows KD as the best method, with DRCA and KD-DRCA performing better in task 2 than task 1. 

\begin{figure*}[!ht]
    \centering
    \includegraphics[width=0.9\linewidth]{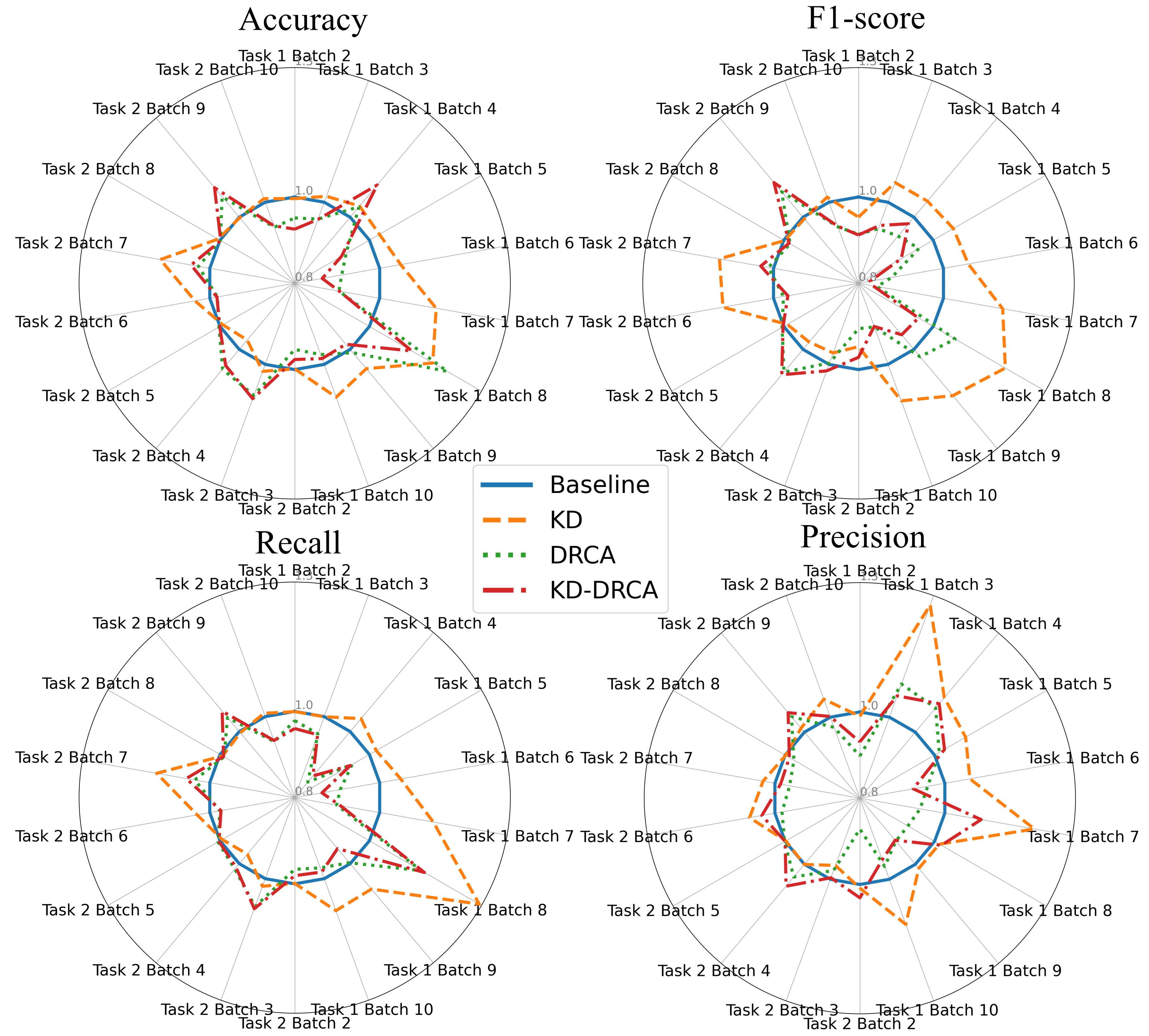}
    \caption{\textbf{Summary radar plots of the mean classification performance of different methods relative to baseline across four metrics for all tasks on the test sets.} The radar plots compare four methods: Baseline, KD, DRCA, and KD-DRCA, across two tasks and nine target batches for each task. Each axis represents a task-batch combination, with values showing the median performance over 30 parallel experiments (normalized to baseline). Task 1: Train on the first batch and test on the other batches; Task 2: Train on all previous batches and test on the next batch.}
    \label{fig:radar}
\end{figure*}

Radar plots in Fig. \ref{fig:radar} show the mean test set performance across metrics, batches, and tasks. Fig. S2 shows validation results, while Fig. S3 and Fig. S4 present the median values on test and validation sets, respectively.
Overall, the superior classification performance from KD is both more frequent and to a higher degree, and KD significantly outperforms the previously validated DRCA method \cite{zhang2017anti}, making it a more suitable choice for electronic nose gas classification drift compensation.

\section{Discussion}
This study systematically evaluated the effectiveness of the KD and DRCA methods, and their combination, to address sensor drift in electronic-nose-based gas classification, a field where accurate gas classification is highly significant but often challenged by sensor drift\cite{vergara2012chemical}. Although prior studies confirmed the effectiveness of certain methods on this public dataset, they often used only single trials and lacks statistical rigor. Our primary novelty of this study lies in designing a statistically validated experimental setup with robust cross-domain testing. This setup addresses gaps in prior studies, which often lacked randomization and rigorous statistical validation. Additionally, we implemented the novel KD method, which is applied for the first time to drift compensation in electronic-nose systems\cite{alkhulaifi2021knowledge,gou2021knowledge} — using the previously validated DRCA as a benchmark \cite{zhang2017anti}. Most importantly, under the systematic and statistical validation, KD achieves the best performance for electronic nose drift compensation, marking a significant advancement in this research area.

Secondly, previous studies, such as Zhang et al. \cite{zhang2017anti}, have applied DRCA, but their experimental design was not the most ideal: using batch 2 to train a model predict batch 3, and using batch 3 to train a model to predict batch 4, whereas in real-world scenarios, models typically are trained on all prior data (e.g., batch 1-4) and then used to predict unseen new data (e.g., batch 5)\cite{liang2020simaug}. This type of more realistic drift compensation is likely more challenging because combining all previous data batches could introduce heterogeneous domain differences, making predictions across these aggregated domains more difficult than predictions between single batches. Therefore, DRCA needs to be validated again for a more realistic setting. Additionally, they did not conduct parallel experiments to verify model robustness, nor did they perform statistical significance testing. Our study sought to fill these gaps by testing these methods across multiple cross-domain prediction tasks with rigorous statistical validation.

Additionally, we explore a combined approach using KD and DRCA. The hybrid KD-DRCA method was designed to leverage the strengths of both techniques: DRCA was used to minimize domain differences\cite{wang2021unsupervised}, and KD was employed to enhance model generalization by transferring knowledge from a complex teacher model to a simpler student model\cite{abbasi2020modeling}. However, the results indicated that KD alone outperformed both DRCA and the KD-DRCA hybrid, making it the most suitable method for mitigating sensor drift in gas sensor applications. Compared to the data-perspective methods like DRCA, KD retains original knowledge while adapting to new target data, potentially preserving more stable \( X \rightarrow Y \) relationship information across domains. In contrast, DRCA, which modifies the data through projection before modeling, might be more prone to overfitting noise information between domains. This finding highlights KD's effectiveness in scenarios where sensor drift is not as pronounced, as its semi-supervised nature appears to be more adept at handling the nuances of the drift problem\cite{gomes2022survey}. Despite overall good results achieved by KD, no drift compensation method is guaranteed to improve the performance under the sensor drift in all experiments, which calls for better control in the application of electronic nose to begin with besides drift compensation.

Despite the promising results, there are limitations to this study: our study did not include a comparison with more complicated methods such as Cycle-GAN\cite{zhu2017unpaired} or reliability-based semi-supervised learning approaches\cite{liu2021boost} that have shown promise in drift compensation. Future research may benefit from incorporating these methods to for a more comprehensive evaluation. Furthermore, the current study was limited to a single dataset. Therefore, in the future, validating the findings across other datasets could strengthen the generalizability of the results.

%%%%%%%%%%%%%%%%%%%%%%%%%%%%%%
\section{Conclusion}
This study aimed to improve the effectiveness of domain adaptation in mitigating sensor drift in electronic-nose-based gas classification. By evaluating the KD, DRCA, and KD-DRCA methods, our results demonstrated that the KD method consistently outperformed both DRCA and KD-DRCA across various metrics, including accuracy, F1-score, precision, and recall. Notably, this is the first application of KD to sensor drift compensation, highlighting its potential as a superior strategy for addressing drift in gas sensors. Despite this, KD-DRCA performed no worse than DRCA, suggesting that combining strategies can still offer some benefits under specific conditions. This study provides a robust foundation for the continued development and optimization of sensor drift compensation in electronic-nose applications, ultimately contributing to more robust gas classification systems in real-world environments.

\bibliographystyle{IEEEtran}
\bibliography{cited}

\end{document}